\documentclass[journal]{IEEEtran}
\usepackage{amsmath,amsfonts}
\usepackage{algorithmic}
\usepackage{algorithm}
\usepackage{array}
\usepackage[caption=false,font=normalsize,labelfont=sf,textfont=sf]{subfig}
\usepackage{textcomp}
\usepackage{stfloats}
\usepackage{url}
\usepackage{verbatim}
\usepackage{graphicx}
\usepackage{cite}
\usepackage{multirow}
\usepackage{booktabs}
\usepackage[table]{xcolor}
\usepackage{booktabs}

\begin{document}

\title{Beyond Fixed Inference: Quantitative Flow Matching for Adaptive Image Denoising}

\author{Jigang Duan, Genwei Ma$^\star$,~\IEEEmembership{Member,~IEEE}, Xu Jiang, Wenfeng Xu, Ping Yang and Xing Zhao$^\star$
\thanks{This work was supported by the National Natural Science Foundation of China (Grant No. 12426308) and the Beijing High Innovation Plan ("Capital High-End Leading Talents Aggregation and Cultivation Program"), Grant No. 202504841094. \textit{(Jigang Duan and
Genwei Ma contributed equally to this work.)} \textit{(Corresponding authors: Genwei Ma and Xing Zhao.)}}
\thanks{Jigang Duan, Xu Jiang, Wenfeng Xu and Xing Zhao are with the School of Mathematical Sciences, Capital Normal University, Beijing 100048, China (e-mail: hiduanjigang@163.com; chianghsu97@gmail.com; willfore@163.com; zhaoxing\_1999@126.com).}
\thanks{Genwei Ma is with the National Center for Applied Mathematics Beijing, Capital Normal University and Academy for multidisciplinary studies, Capital Normal University, Beijing, 100048, China (e-mail: magenwei@126.com).}
\thanks{Ping Yang is with the Institute of Nuclear and New Energy Technology, Tsinghua University, Beijing 100084, China (e-mail: yang\_ping0603@163.com).}}

\markboth{Journal of \LaTeX\ Class Files, April~2026}%
{Shell \MakeLowercase{\textit{et al.}}: A Sample Article Using IEEEtran.cls for IEEE Journals}

\maketitle

\begin{abstract}
  Diffusion and flow-based generative models have shown strong potential for image restoration. However, image denoising under unknown and varying noise conditions remains challenging, because the learned vector fields may become inconsistent across different noise levels, leading to degraded restoration quality under mismatch between training and inference. To address this issue, we propose a quantitative flow matching framework for adaptive image denoising. The method first estimates the input noise level from local pixel statistics, and then uses this quantitative estimate to adapt the inference trajectory, including the starting point, the number of integration steps, and the step-size schedule. In this way, the denoising process is better aligned with the actual corruption level of each input, reducing unnecessary computation for lightly corrupted images while providing sufficient refinement for heavily degraded ones. By coupling quantitative noise estimation with noise-adaptive flow inference, the proposed method improves both restoration accuracy and inference efficiency. Extensive experiments on natural, medical, and microscopy images demonstrate its robustness and strong generalization across diverse noise levels and imaging conditions.
\end{abstract}


\section{Introduction}

\IEEEPARstart{I}{mage} denoising is a fundamental problem in low-level vision. Its goal is to improve visual quality while providing cleaner and more reliable inputs for downstream high-level vision tasks, such as detection and segmentation, as well as for quantitative analysis. In practical applications, however, image noise is rarely homogeneous or idealized. Its magnitude and statistical properties often vary across space and are closely coupled with the imaging device, acquisition process, and measurement conditions. For example, in natural images, noise characteristics can vary substantially across camera sensors and imaging pipelines \cite{zou2025calibrationfree}. In computed tomography (CT), acquisition settings such as tube voltage and tube current, together with detector technology, strongly affect image noise, contrast-to-noise performance, and dose efficiency \cite{rohme2024abdominalct,vanderbie2025pcctreview}. In electron microscopy, the effective noise behavior is closely related to electron dose, specimen sensitivity, and detector capability, which directly influence structural interpretation and quantitative analysis \cite{kim2025shine,zhan2024lowdoseem}. Such heterogeneity makes it difficult to develop denoising methods that generalize reliably across imaging conditions and noise regimes.

\begin{figure}[htbp]
  \centering
  \includegraphics[width=1.0\linewidth]{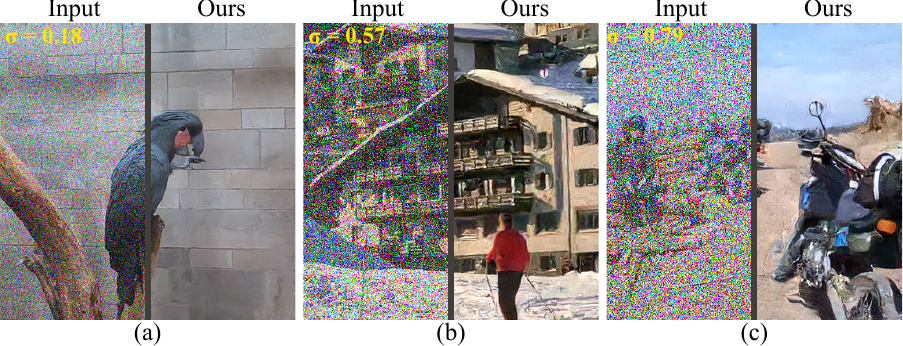}
  \caption{Qualitative denoising results of the proposed method under different Gaussian noise levels. For each example, the left image is the noisy input and the right image is the corresponding denoised result. The noise standard deviation $\sigma$ is shown in the upper-left corner of each noisy image. (a)--(c) correspond to medium-, high-, and very high-noise conditions, respectively.}
  \label{fig:compare}
\end{figure}

Existing denoising methods can generally be divided into model-driven and data-driven approaches. Classical model-driven methods rely on explicit degradation assumptions and hand-crafted image priors, and typically formulate denoising as a regularized optimization problem. Representative examples include variational methods, especially total-variation-based formulations \cite{Beck2009TIP}; sparse representation and dictionary learning methods such as K-SVD \cite{EladAharon2006TIP}; non-local self-similarity methods including NLM and BM3D/CBM3D \cite{Buades2005CVPR,Dabov2007BM3D_TIP,Dabov2007CBM3D_ICIP}; and low-rank modeling methods \cite{Gu2014WNNM_CVPR}. Although effective under specific assumptions, these methods depend heavily on the validity of the prescribed priors and noise models. When the noise is spatially non-uniform, signal-dependent, or dynamically varying, fixed prior models often lack sufficient adaptability, leading to noticeable performance degradation on real captured images \cite{Plotz2017DND_CVPR}.

Deep learning has substantially advanced image denoising, with modern neural networks often outperforming classical methods across a wide range of benchmarks and imaging scenarios \cite{zhang2017dncnn,guo2019cbdnet,zamir2022restormer,chen2022nafnet}. Existing deep denoising methods can be broadly divided into supervised and unsupervised/self-supervised paradigms. Supervised methods learn restoration mappings from paired noisy-clean data and can achieve impressive performance, but collecting high-quality paired data is often expensive or impractical, and their generalization strongly depends on the coverage of noise types and levels in the training set, especially when transferring from synthetic to real-world noise \cite{abdelhamed2018sidd}. In contrast, unsupervised and self-supervised methods reduce the reliance on clean labels by constructing supervision directly from noisy observations \cite{lehtinen2018noise2noise,krull2019noise2void,batson2019noise2self,huang2021neighbor2neighbor}. However, their effectiveness usually relies on assumptions such as noise independence, blind-spot masking, or local neighborhood constraints. When these assumptions are violated, especially under correlated noise, structured perturbations, or strong signal-noise coupling, the restored results may suffer from oversmoothing, detail loss, or spurious structures \cite{krull2019noise2void,kim2021noise2score,chihaoui2024mash}.

Recent advances in generative modeling have opened a promising direction for image denoising by moving beyond direct deterministic regression toward conditional distribution modeling \cite{autodir2024eccv,spire2024eccv,dualdn2024eccv,garber2024cvpr_ipg,ding2024cvpr_constrained}. Among these approaches, diffusion models and flow-matching methods have emerged as two representative paradigms. Diffusion models often provide strong restoration quality, particularly in texture recovery and edge preservation, but they typically require multiple iterative sampling steps to produce high-fidelity results \cite{lin2024diffbir,ye2024dtpm,conde2024instructir,chen2025unirestore}. By contrast, flow matching and rectified flow formulate generation or restoration through learned transport trajectories, making them attractive for low-step inference and flexible numerical solvers \cite{resflow2024icml,yan2024perflow,lee2024neurips_rfpp,kornilov2024neurips_ofm,stoica2025contrastive}. To reduce the computational cost of iterative generative inference, recent studies have explored diffusion distillation and few-step sampling \cite{mei2024codi,sauer2024add,garber2025cm4ir}, as well as improved trajectory design, rectified-flow training, and transport optimization for efficient low-step flow-based inference \cite{cheng2025c2ot,ke2025proreflow,chadebec2025lbm}. However, such acceleration strategies often become more sensitive to mismatches between noise conditions and inference configurations, which may lead to oversmoothing, detail loss, or spurious textures \cite{dmd2024cvpr,zhou2024neurips_sfd,luo2024neurips_sim,xie2024neurips_emd,bird2024neurips}. In particular, flow-matching-based denoising formulates restoration as the numerical integration of an ordinary differential equation (ODE) along a learned time-dependent vector field that transports a noisy state toward a clean image. Although this formulation is naturally compatible with higher-order solvers and few-step inference, it still faces a fundamental trade-off between restoration fidelity and computational efficiency. Achieving high-quality denoising with only a small number of integration steps across diverse noise levels therefore remains challenging.

More importantly, diffusion- and flow-based denoising methods face a practically important challenge under varying noise levels: a fixed inference configuration is often not equally suitable for inputs from different noise regimes. As a result, the same setting may be computationally redundant for relatively clean inputs, while being insufficient for heavily corrupted ones. A natural solution is therefore to estimate the input noise level before inference and adapt the inference process accordingly. Motivated by this observation, we propose a novel adaptive denoising framework, termed \emph{Quantitative Flow Matching} (QFM). Specifically, we first estimate the global noise level of the input image using a simple yet effective statistic computed from local pixel differences over non-overlapping $2\times2$ blocks. By exploiting the local smoothness prior of natural images, the estimator suppresses image structures while enhancing noise-related fluctuations, yielding a robust estimate of the overall noise level. We then map the estimated noise level to an appropriate location on the flow trajectory and start the integration from this adaptive point instead of using a fixed initialization. In addition, both the total number of integration steps and the step-size schedule are adjusted according to the estimated noise level, enabling faster progression under heavier noise and finer refinement near cleaner stages of the trajectory. In this way, the proposed method reduces redundant function evaluations while maintaining high denoising fidelity. Qualitative examples under different noise levels are shown in Fig.~\ref{fig:compare}, indicating that the proposed method performs favorably across substantially different noise regimes and remains effective even under severe noise conditions.

The main contributions and advantages of the proposed method are summarized as follows:

\begin{itemize}
  \item We systematically analyze the ambiguity and inconsistency of learned vector fields in existing diffusion- and flow-based denoising methods under varying noise levels and distributions, providing a principled explanation for their performance degradation under out-of-distribution noise conditions.

  \item We propose a novel quantitative flow-matching denoising framework that adapts the inference process according to an estimated noise level. By configuring the starting point, the number of integration steps, and the step-size schedule in a noise-aware manner, the proposed method achieves both efficient and high-quality denoising.

  \item We introduce a simple yet effective noise estimation method based on local pixel variations. By exploiting the local smoothness prior of images, it estimates the global noise level from statistics computed over non-overlapping $2\times2$ image blocks.

  \item We conduct extensive experiments on both natural images and medical images, including CT and electron microscopy data, to validate the effectiveness of the proposed method, demonstrating strong robustness and competitive performance across diverse noise environments.

  \item The proposed framework improves the practical applicability of diffusion- and flow-based denoising in non-ideal noise scenarios, providing a useful solution for real-world image restoration tasks.
\end{itemize}

The remainder of this paper is organized as follows. Section \ref{sec:related_theory} analyzes the limitations of existing methods and the motivation for the proposed approach. Section \ref{sec:method} presents the proposed method, including noise estimation and adaptive flow-matching inference. Section \ref{sec:experiments} reports the experimental results, and Section \ref{sec:conclusion} concludes this paper.

\section{Related Theory and Motivation}
\label{sec:related_theory}

\begin{figure*}[ht]
  \centering
  \includegraphics[width=1.0\linewidth]{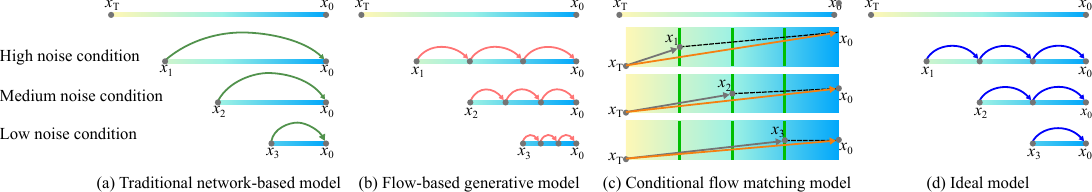}
  \caption{Conceptual comparison of denoising mechanisms under different noise conditions. (a) Traditional network-based model. (b) Flow-based generative model with a fixed inference trajectory. (c) Conditional flow matching model guided by the noisy observation. (d) Ideal noise-aware model with trajectories better matched to different noise levels. Here, $x_0$ denotes the clean image, $x_T$ denotes the noisy initial state, and $x_1$, $x_2$, and $x_3$ represent inputs under high-, medium-, and low-noise conditions, respectively.}
  \label{fig:method_compare}
\end{figure*}

To motivate the proposed method, we first analyze the theoretical basis and practical limitations of existing deep learning-based denoising methods, with particular emphasis on generative-model-based approaches. Among generative methods, we use Flow Matching as a representative example, while noting that the resulting insights are also relevant to related frameworks such as diffusion models. Fig.~\ref{fig:method_compare} provides a conceptual comparison of the denoising mechanisms underlying different categories of methods. Based on this analysis, we then examine their limitations under varying noise levels and distributions, which leads to the proposed Quantitative Flow Matching framework for adaptive denoising under such conditions.

\subsection{Traditional Network-Based Models}

Conventional network-based methods typically learn either a direct mapping from noisy images to clean images \cite{herbreteau2025noise2vst} or, in unsupervised settings, a denoising function from supervisory signals derived from noisy observations \cite{liao2024amsnet,li2025positive2negative}. Their performance depends strongly on how well the noise characteristics represented in the training data match those encountered at test time, particularly in terms of noise type and level \cite{kim2025idf,ma2025pixel2pixel}. When the noise level or distribution of a test image deviates substantially from those seen during training, the model may generalize poorly and produce degraded restoration results, often manifesting as oversmoothing, loss of fine details, or visible artifacts \cite{kim2024lan,cheng2024tclip,herbreteau2025noise2vst}.

As illustrated in Fig.~\ref{fig:method_compare}(a), $x_T$ denotes the noisy input state and $x_0$ denotes the clean image, while intermediate cases such as $x_1$, $x_2$, and $x_3$ represent high-, medium-, and low-noise conditions, respectively. In general, supervised and unsupervised denoising methods perform most reliably when the test-time noise characteristics are close to those represented in the training data. Their performance, however, may degrade substantially when the input noise level is markedly higher or lower than the training regime. As will be shown in the experiments, such methods remain limited in their ability to generalize across a broad range of noise levels when test-time noise characteristics differ substantially from those seen during training.

\subsection{Flow-Based Generative Models: A Case Study of Flow Matching}

Let \(q(x_0)\) denote the data distribution. Flow Matching constructs a time-dependent probability path \(p_t(x)\), with \(t \in [0,1]\), that continuously transforms a simple prior distribution \(\pi(x)\) into the data distribution, such that \(p_1(x)\approx \pi(x)\) and \(p_0(x)=q(x_0)\). The path \(p_t(x)\) is generated by a time-dependent vector field \(u_t(x)\) satisfying the continuity equation
\begin{equation}
  \frac{\partial p_t(x)}{\partial t} = - \nabla \cdot \bigl(p_t(x) u_t(x)\bigr).
\end{equation}
A common construction introduces a conditional probability path \(p_t(x \mid x_0, x_T)\), where \(x_0 \sim q(x_0)\) and \(x_T \sim \pi(x)\). For example, under a linear interpolation path,
\begin{equation}
  \begin{split}
    &p_t(x \mid x_0, x_T) = \mathcal{N}\!\bigl(x \mid \mu_t(x_0,x_T), \sigma_t^2 I\bigr), \\
    &\mu_t = t x_T + (1-t)x_0.
  \end{split}
\end{equation}
The corresponding conditional vector field is
\begin{equation}
  u_t(x \mid x_0, x_T) = x_0 - x_T.
\end{equation}
The Flow Matching (FM) objective is then written as
\begin{equation}
  \mathcal{L}_{\mathrm{FM}}(\theta)
  =
  \mathbb{E}_{t, x_0, x_T}
  \Bigl[
    \bigl\| v_\theta(x_t,t) - u_t(x_t \mid x_0,x_T) \bigr\|^2
  \Bigr],
\end{equation}
where \(t \sim \mathcal{U}[0,1]\), \(x_0 \sim q(x_0)\), \(x_T \sim \pi(x)\), and \(x_t \sim p_t(x \mid x_0,x_T)\). For commonly used paths such as independent coupling, this objective reduces to a simpler regression form,
\begin{equation}
  \mathcal{L}_{\mathrm{FM}}(\theta)
  =
  \mathbb{E}_{t, x_0, x_T}
  \Bigl[
    \bigl\| v_\theta(x_t,t) - (x_0 - x_T) \bigr\|^2
  \Bigr].
\end{equation}

After training, the learned vector field \(v_\theta(x,t)\) approximates the generating field, and sampling is performed by solving the associated ordinary differential equation backward in time,
\begin{equation}
  \frac{dx}{dt} = v_\theta(x,t),
  \qquad x_T\sim \pi(x),
\end{equation}
from \(t=1\) to \(t=0\). The resulting solution \(x(0)\) is then a sample from the learned distribution \(p_0^\theta(x)\approx q(x)\).

While flow matching provides a principled framework for image generation by transporting pure noise to clean data, its direct use for general image denoising, where the input may exhibit arbitrary and unknown noise levels and distributions, introduces a fundamental challenge. Recent flow-based restoration and inverse-problem formulations have shown that conditioning ambiguity and restoration uncertainty become central issues once the observation no longer corresponds to a fixed, uniquely specified degradation setting \cite{qin2025resflow,meanti2025ddm,kim2025flowdps}.

Consider the standard denoising training setting illustrated in Fig.~\ref{fig:method_compare}(b). Let \(x_0\) denote a clean image, and let \(x_2\) denote its noisy observation under a fixed and homogeneous noise condition. The conditional interpolation path is defined by
\begin{equation}
  x_t = t x_2 + (1-t)x_0,
\end{equation}
with the corresponding target vector field
\begin{equation}
  u_t(x_t \mid x_0,x_2) = x_0 - x_2.
\end{equation}
A neural network \(v_\theta(x_t,t)\) is trained by minimizing
\begin{equation}
  \mathcal{L}_{\mathrm{FM}}
  =
  \mathbb{E}_{t,x_0,x_2}
  \bigl\|
  v_\theta(x_t,t) - (x_0-x_2)
  \bigr\|^2.
\end{equation}
Under such a homogeneous training regime, the displacement target associated with each noisy observation remains consistent, which makes the learning problem well defined.

The difficulty arises when the model is applied to denoising under mismatched noise conditions. As illustrated in Fig.~\ref{fig:method_compare}(b), let \(x_1\) and \(x_3\) denote two noisy versions of the same clean image \(x_0\), corresponding to higher- and lower-noise conditions than the training observation \(x_2\), respectively. Their associated conditional vector fields are
\begin{equation}
  u_t(\cdot \mid x_0,x_1) = x_0 - x_1,
  \qquad
  u_t(\cdot \mid x_0,x_3) = x_0 - x_3.
\end{equation}
Since these noisy observations may generate visually similar intermediate states along different interpolation paths, there may exist \(t,s \in [0,1]\) such that
\begin{equation}
  x_t = t x_1 + (1-t)x_0 = s x_3 + (1-s)x_0.
\end{equation}
At such an intermediate state, the target displacement is no longer uniquely determined by \((x_t,t)\), because multiple valid vector-field targets may exist under different noise conditions. In other words, unless the underlying noise condition is explicitly encoded, a deterministic network parameterization \(v_\theta(x_t,t)\) is forced to fit an ambiguous regression target. This ambiguity makes the learning objective less well defined under varying or out-of-distribution noise conditions.

Moreover, standard Flow Matching inference typically uses a fixed initialization, a fixed integration horizon, and a fixed discretization strategy. Such a configuration cannot adapt to the varying magnitude of the denoising transformation required by different inputs. In particular, inputs under heavier noise generally require a larger overall displacement toward the clean image than those under lighter noise. A fixed inference trajectory, combined with a vector field learned under noise mismatch, is therefore unlikely to provide an equally suitable restoration process across diverse noise conditions. Recent low-step flow- and bridge-based restoration methods have likewise emphasized that trajectory design, inference scheduling, and conditioning strategy play a critical role in both restoration fidelity and computational efficiency \cite{xu2025flowsr,chadebec2025lbm,ke2025proreflow}. These observations motivate the development of a noise-aware and quantitatively adaptive denoising framework.

\subsection{Conditional Flow Matching Model}

Conditional flow matching (CFM) aims to model the conditional distribution \(q(x_0 \mid y)\) given an observed condition \(y\). In this setting, the initial distribution becomes a condition-dependent prior \(\pi(x \mid y)\), and both the probability path and the generating vector field are conditioned on \(y\). A conditional path can therefore be constructed between a prior sample \(x_T \sim \pi(x \mid y)\) and the target sample \(x_0 \sim q(x_0 \mid y)\), with the corresponding conditional vector field driving the evolution of \(x_t\) under the condition \(y\) \cite{meanti2025ddm,kim2025flowdps}.

The training objective of CFM can be written as
\begin{equation}
  \mathcal{L}_{\mathrm{CFM}}(\theta)
  =
  \mathbb{E}_{t, y, x_0, x_T}
  \left[
    \left\|
    v_\theta(x_t, t, y) - u_t(x_t \mid x_0, x_T, y)
    \right\|^2
  \right],
\end{equation}
where \(x_0 \sim q(x_0 \mid y)\), \(x_T \sim \pi(x \mid y)\), \(x_t\) is sampled from the corresponding conditional path, and \(t \sim \mathcal{U}[0,1]\). Compared with unconditional Flow Matching, the model \(v_\theta\) now takes the observed noisy image \(y\) as an additional condition. Sampling from the learned conditional distribution is performed by integrating
\begin{equation}
    \frac{dx}{dt} = v_\theta(x,t,y),
\qquad x(1)\sim \pi(x \mid y),
\
\end{equation}
from \(t=1\) to \(t=0\).

In conditional flow matching for denoising, the generation process is explicitly guided by the noisy observation \(y\). Under an idealized linear path, the learned vector field can be interpreted conceptually as being influenced by two factors, as conceptualized in Fig.~\ref{fig:method_compare}(c): a global transport tendency toward the clean target and a condition-dependent guidance signal extracted from the noisy input \(y\). The latter is typically injected into \(v_\theta(x,t,y)\) through conditioning mechanisms such as feature modulation or cross-attention. The effectiveness of this guidance therefore depends strongly on whether the noise characteristics of \(y\) are consistent with those represented during training \cite{kim2025idf,ma2025pixel2pixel}.

Let \(\sigma_y\) denote the noise level of the conditional input \(y\), and let \(\Sigma_{\mathrm{train}}\) denote the noise regime represented during training. When \(\sigma_y\) deviates substantially from \(\Sigma_{\mathrm{train}}\), the guidance extracted from \(y\) may no longer be fully compatible with the behavior learned by the model. In the high-noise mismatch case (\(\sigma_y \gg \Sigma_{\mathrm{train}}\)), the conditional input is dominated more strongly by noise, so the extracted guidance may contain misleading structural cues, causing the inferred trajectory to deviate markedly from the desired path toward the clean image. This often leads to severe artifacts or large restoration errors. In the low-noise mismatch case (\(\sigma_y \ll \Sigma_{\mathrm{train}}\)), the conditional input contains weaker corruption than the regime emphasized during training. As a result, the inferred trajectory may apply unnecessarily strong denoising behavior, which can suppress fine structures and produce oversmoothed results. Although the deviation is often less severe than in the high-noise case, the restoration may still be suboptimal in terms of high-frequency fidelity.

\begin{figure*}[ht]
  \centering
  \includegraphics[width=0.95\linewidth]{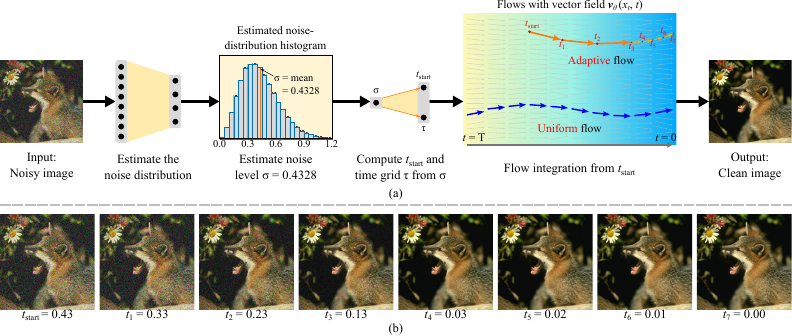}
  \caption{Overview of the proposed method. (a) Overall pipeline, including noise-level estimation, computation of the adaptive starting time and inference grid, and reverse flow integration for denoising. (b) Intermediate denoising results at different time points during adaptive inference.}
  \label{fig:flow_chart}
\end{figure*}

\subsection{Motivation for Quantitative Flow-Based Model Denoising}

The above analysis suggests that standard flow-matching-based denoising methods depend strongly on the noise regime represented during training. When the test-time noise characteristics deviate substantially from this regime, both the learned guidance and the fixed inference configuration may become suboptimal, resulting in degraded restoration quality or unnecessary computational cost. This observation motivates an adaptive framework that explicitly incorporates a quantitative estimate of the input noise level into the inference process. By estimating the noise level before inference, the starting point of integration, the total number of steps, and the step-size schedule can be adjusted to better match the actual corruption level of the input image. Such a strategy improves both denoising fidelity and computational efficiency across a wide range of noise conditions, as conceptually illustrated in Fig.~\ref{fig:method_compare}(d).

\section{Method}
\label{sec:method}

In this section, we present the proposed denoising framework, which combines quantitative noise estimation with adaptive flow-matching inference. The overall pipeline is shown in Fig.~\ref{fig:flow_chart}(a). Given a noisy input image, the method first estimates its noise level, which is then used to determine the starting point of flow-matching inference on the normalized time axis. The image is subsequently refined by integrating along the learned vector field using a coarse-to-fine step-size schedule. Example intermediate denoising results are shown in Fig.~\ref{fig:flow_chart}(b). The two main components of the proposed framework, namely quantitative noise estimation and adaptive flow-matching inference, are detailed below.

\subsection{Quantitative Noise Estimation}

Given a noisy image \(Y\in\mathbb{R}^{H\times W}\), we model it as a clean image \(X\) corrupted by noise. The proposed estimator is applicable to both additive Gaussian noise and signal-dependent Poisson noise. For clarity, we first present the derivation under additive Gaussian noise:
\begin{equation}
  Y(n)=X(n)+\epsilon(n),\qquad \epsilon(n)\stackrel{\mathrm{iid}}{\sim}\mathcal{N}(0,\sigma^2).
  \label{eq:noise_model}
\end{equation}
Here, \(n\) denotes the pixel index and \(\sigma^2\) is the global noise variance. Our goal is to estimate the global noise level from a single observation \(Y\), without relying on training data.

\begin{figure*}[ht]
  \centering
  \includegraphics[width=0.95\linewidth]{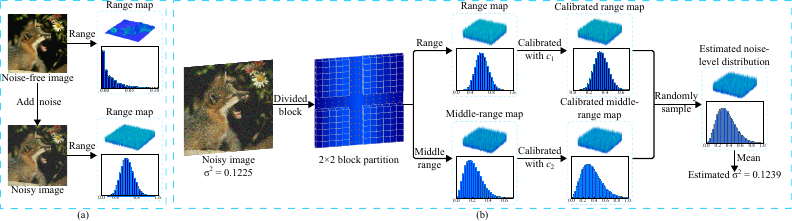}
  \caption{Pipeline of the proposed quantitative noise estimation method. (a) Illustration of block-wise pixel differences in noise-free and noisy images after partitioning into non-overlapping $2\times2$ patches. (b) Estimation of the global noise level using range and middle-range statistics, followed by calibration, random fusion, and sample averaging.}
  \label{fig:esti_noise}
\end{figure*}

Based on the local smoothness prior of natural images, we assume that the clean signal \(X\) is approximately constant within a sufficiently small neighborhood. As illustrated in Fig.~\ref{fig:esti_noise}(a), when a noise-free image is partitioned into non-overlapping \(2\times2\) blocks, the within-block pixel differences are typically small, and the corresponding range values are concentrated near zero. In contrast, applying the same operation to a noisy image produces much stronger random fluctuations.

The estimation pipeline is shown in Fig.~\ref{fig:esti_noise}(b). We first partition \(Y\) into non-overlapping \(2\times2\) blocks \(\{B_k\}_{k=1}^{K}\), where \(K=\lfloor H/2\rfloor\lfloor W/2\rfloor\). For each block \(B_k\), its four pixel values are sorted in ascending order as
\[
  Y^{(k)}_{(1)}\le Y^{(k)}_{(2)}\le Y^{(k)}_{(3)}\le Y^{(k)}_{(4)},
\]
and two local statistics are defined as
\begin{equation}
  \begin{aligned}
    \text{Range:}\quad D_{1,k} &= Y^{(k)}_{(4)}-Y^{(k)}_{(1)},\\
    \text{Middle range:}\quad D_{2,k} &= Y^{(k)}_{(3)}-Y^{(k)}_{(2)}.
  \end{aligned}
\end{equation}
The range \(D_{1,k}\) is more sensitive to noise in flat regions, but it can also be affected by signal variation near edges. By contrast, the middle range \(D_{2,k}\) is less sensitive to extreme values and is therefore generally more robust in structured regions.

Under the local-constant assumption, the pixels within a block can be written as
\[
  Y_i=x+\sigma Z_i,\qquad Z_i\stackrel{\mathrm{iid}}{\sim}\mathcal{N}(0,1).
\]
Therefore, for any block \(k\),
\begin{equation}
  D_{1,k}=\sigma U_1,\qquad D_{2,k}=\sigma U_2,
  \label{eq:Di_sigmaUi}
\end{equation}
where \(U_1=Z_{(4)}-Z_{(1)}\) and \(U_2=Z_{(3)}-Z_{(2)}\) depend only on the order statistics of the standard normal distribution. It follows that
\begin{equation}
  \mathbb{E}[D_{1,k}]=c_1\sigma,\qquad \mathbb{E}[D_{2,k}]=c_2\sigma,
  \label{eq:Di_expect}
\end{equation}
where \(c_1=\mathbb{E}[U_1]\approx 2.06\) and \(c_2=\mathbb{E}[U_2]\approx 0.59\).

Accordingly, for each \(2\times2\) block \(B_k\), we construct two block-wise noise estimates as
\begin{equation}
  S_{1,k}=\frac{D_{1,k}}{c_1},\qquad
  S_{2,k}=\frac{D_{2,k}}{c_2}.
  \label{eq:block_sigma_map}
\end{equation}
Using the range alone performs well in flat regions but may be biased near edges, whereas the middle range is more robust but statistically less efficient. To balance robustness and statistical efficiency, we fuse the two estimates through random partitioning. Specifically, the index set \(\{1,\dots,K\}\) is randomly divided into two disjoint subsets \(\mathcal{I}_1\) and \(\mathcal{I}_2\) such that
\begin{equation}
  \mathcal{I}_1\cap \mathcal{I}_2=\emptyset,\qquad
  \mathcal{I}_1\cup \mathcal{I}_2=\{1,\dots,K\},
\end{equation}
with the size ratio
\[
  |\mathcal{I}_1|:|\mathcal{I}_2|=1:4.
\]
Under a fixed total number of blocks, this ratio balances the statistical uncertainty contributed by the two estimators. The range-based estimator is approximately four times as statistically efficient as the middle-range-based estimator. For each \(k\in\{1,\dots,K\}\), the fused estimate is defined as
\begin{equation}
  S_k=
  \begin{cases}
    S_{1,k}, & k\in\mathcal{I}_1,\\
    S_{2,k}, & k\in\mathcal{I}_2.
  \end{cases}
  \label{eq:fused_Sk}
\end{equation}

Finally, the sample mean of the fused block-wise estimates is taken as the global noise standard deviation:
\begin{equation}
  \hat{\sigma}=\frac{1}{K}\sum_{k=1}^{K} S_k.
  \label{eq:sigma_final}
\end{equation}
The corresponding estimate of the global noise variance is then given by \(\hat{\sigma}^2\).

\subsection{Quantitative Flow Matching with Normalized Vector Fields}

After obtaining the global noise estimate \(\hat{\sigma}\), we perform flow-matching inference on a normalized time axis. A pretrained flow-matching model is used to predict a time-dependent vector field, and the denoised image is recovered by integrating along this field from a noise-dependent starting time toward the clean endpoint.

We define a normalized time variable \(t\in[0,1]\), where \(t=1\) corresponds to the maximally noisy state \(x_T\) and \(t=0\) corresponds to the clean image \(x_0\). In our framework, \(x_T\) represents the image at the highest noise level, characterized by a standard deviation \(\sigma\). During training, we are given a clean image \(x_0\) and its noisy observation \(x_1\), where the noise standard deviation of \(x_1\) is a lower and known value \(\hat{\sigma}\) satisfying \(\hat{\sigma}\leq \sigma\). To construct a normalized trajectory that generalizes across noise levels, we relate \(x_1\) to the normalized noisy endpoint \(x_T\) under a linear scaling assumption anchored at \(x_0\):
\begin{equation}
  x_T = x_0 + \sigma \epsilon, \qquad x_1 = x_0 + \hat{\sigma}\epsilon,
\end{equation}
where \(\epsilon\sim\mathcal{N}(0,I)\) is a shared noise vector. It then follows that \(x_1\) and \(x_T\) are collinear with respect to \(x_0\) in the noise space, which gives
\begin{equation}
  x_T = x_0 + \frac{\sigma}{\hat{\sigma}}(x_1-x_0).
\end{equation}

Based on this relation, we define an interpolation path between \(x_0\) and \(x_1\) as
\begin{equation}
  x_t = (1-t)x_0 + tx_1, \qquad t\in[0,1],
  \label{eq:norm_path}
\end{equation}
and write the corresponding normalized target vector field as
\begin{equation}
  u_t(x_t)=\frac{dx_t}{dt}=x_T-x_0=\frac{\sigma}{\hat{\sigma}}(x_1-x_0).
  \label{eq:true_field}
\end{equation}
Compared with the naive direction \((x_1-x_0)\) defined at the specific noise level \(\hat{\sigma}\), this target field introduces a normalization factor \(\sigma/\hat{\sigma}\). Accordingly, we train a neural network \(v_\theta(x,t,\hat{\sigma})\) to approximate this normalized vector field by minimizing
\begin{equation}
  \mathcal{L}(\theta)=\mathbb{E}_{t,x_0,\hat{\sigma},\epsilon}
  \left[
    \left\|
    v_\theta(x_t,t,\hat{\sigma})-\frac{\sigma}{\hat{\sigma}}(x_1-x_0)
    \right\|^2
  \right],
  \label{eq:norm_loss}
\end{equation}
where \(t\sim\mathcal{U}[0,1]\), \(\hat{\sigma}\) is sampled from a noise-level distribution, and \(x_t\) is computed by Eq.~\eqref{eq:norm_path}. By explicitly conditioning on the noise level \(\hat{\sigma}\), the model learns a normalized vector field that adapts to different input noise intensities.

During inference, given a noisy input \(x_1\) with an estimated noise level \(\hat{\sigma}\), we do not explicitly construct the unknown endpoint \(x_T\). Instead, we regard \(x_1\) as lying on the normalized trajectory at a time position determined by \(\hat{\sigma}/\sigma\), and then solve the reverse-time ODE from this starting point to \(t=0\):
\begin{equation}
  \frac{dx}{dt}=v_\theta(x,t),\qquad
  x\!\left(t=\frac{\hat{\sigma}}{\sigma}\right)=x_1
  \longrightarrow
  x(t=0)=\hat{x}_0.
  \label{eq:fm_ode}
\end{equation}
In this way, inputs with different noise levels are embedded into a shared normalized time axis, so that the integration always follows a consistent normalized evolution direction, thereby improving the stability of the denoising process.

To realize noise-adaptive inference, we determine the starting time \(t_{\mathrm{start}}\) according to the estimated noise level. Intuitively, stronger noise corresponds to a starting point closer to \(t=1\), whereas weaker noise corresponds to a starting point closer to \(t=0\). Specifically, the estimated noise standard deviation \(\hat{\sigma}\) is mapped to a start index on a discrete time grid. Let training and inference share a grid of length \(S\), denoted by \(\{t_i\}_{i=0}^{S-1}\), where \(t_0=0\) and \(t_{S-1}=1\). We choose a start index \(i_0\) according to \(\hat{\sigma}\) and set
\begin{equation}
  t_{\mathrm{start}}=t_{i_0},\qquad x_{t_{\mathrm{start}}}=x^{(0)}.
  \label{eq:t0_select}
\end{equation}
In this work, we set \(x^{(0)}=Y\), and integrate from \(t=t_{\mathrm{start}}\) to \(t=0\) to obtain the final output.

After determining \(t_{\mathrm{start}}\), we integrate Eq.~\eqref{eq:fm_ode} over the interval \([t_{\mathrm{start}},0]\). Since the high-noise stage is more tolerant to large step sizes whereas the low-noise stage is more sensitive to integration errors, we adopt a two-level step-size strategy: larger steps are used in the high-noise segment for efficiency, while smaller steps are used near the clean endpoint for stable refinement.

Specifically, inference uses only points from the shared time grid \(\{t_i\}\). Let \(i_{\mathrm{start}}\) satisfy \(t_{i_{\mathrm{start}}}=t_{\mathrm{start}}\), and let \(M\) denote a coarse sampling interval. Along the reverse-time direction, we sample with interval \(M\) to obtain a coarse sequence
\begin{equation}
  \mathcal{T}_{\mathrm{coarse}}
  = \{t_{i_{\mathrm{start}}},\, t_{i_{\mathrm{start}}-M},\, t_{i_{\mathrm{start}}-2M},\, \dots\},
  \label{eq:t_coarse}
\end{equation}
and then fill in the remaining grid points after the last coarse time point (i.e., closer to \(0\)) to obtain a fine sequence \(\mathcal{T}_{\mathrm{fine}}\). The final time grid is defined as
\begin{equation}
  \mathcal{T}=\mathcal{T}_{\mathrm{coarse}}\Vert\mathcal{T}_{\mathrm{fine}}
  =\{t_k\}_{k=0}^{N},\qquad t_0=t_{\mathrm{start}},\; t_N=0,
  \label{eq:t_all}
\end{equation}
where \(\Vert\) denotes concatenation in decreasing-time order, and the adjacent step sizes are \(\Delta t_k=t_k-t_{k+1}>0\).

Given \(\mathcal{T}\), we solve Eq.~\eqref{eq:fm_ode} using the explicit Euler method:
\begin{equation}
  x_{k+1}=x_k-\Delta t_k\,v_\theta(x_k,t_k),\qquad k=0,1,\dots,N-1,
  \label{eq:euler_update}
\end{equation}
where \(x_0=x^{(0)}\), and the final output is \(\hat{X}=x_N\). Since \(t_{\mathrm{start}}\) is determined by \(\hat{\sigma}\) and \(\mathcal{T}\) is adaptively generated on the shared time grid, the inference process automatically adjusts both the reverse integration interval and the step allocation according to the input noise level, thereby reducing computation while preserving reconstruction quality.

This completes the description of the proposed noise-estimation-driven adaptive flow-matching denoising method.

\section{Experiments and Results}
To comprehensively evaluate the proposed normalized quantitative flow-matching denoising model, we conduct systematic experiments on both synthetic and real-world denoising tasks across three application scenarios: synthetic natural-image denoising, fluorescence microscopy denoising, and low-dose CT denoising. This experimental design is intended to validate the generalization capability of the proposed method across different noise types and noise levels. We compare the proposed method with state-of-the-art supervised and unsupervised denoising approaches to demonstrate its performance advantages. In addition, we perform ablation studies to analyze the accuracy of noise estimation and its impact on the overall denoising performance.

\label{sec:experiments}

\subsection{Dataset and Settings}

\begin{figure*}[htbp]
  \centering
  \includegraphics[width=0.95\linewidth]{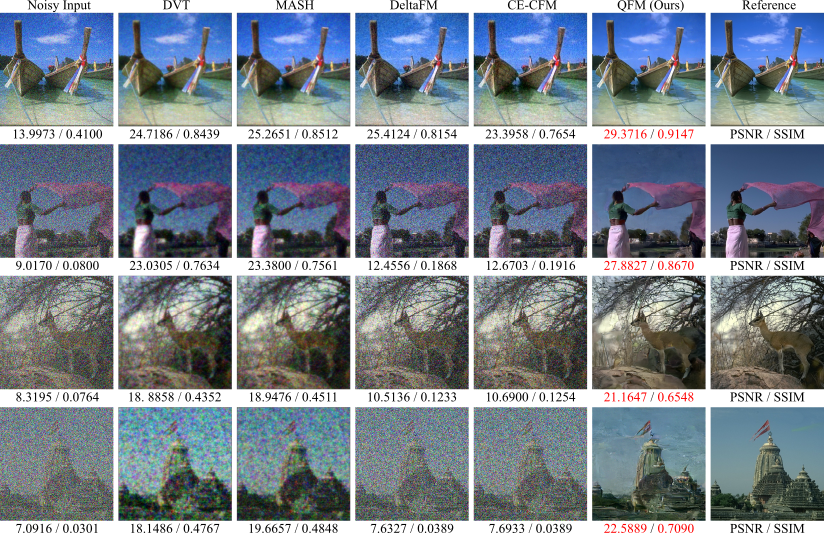}
  \caption{Qualitative denoising comparisons on the BSDS~500 dataset. From left to right: noisy input, supervised baseline DVT \cite{yang2024denoising}, self-supervised baseline MASH \cite{chihaoui2024mash}, flow-based baseline DeltaFM \cite{stoica2025contrastive}, conditional flow-matching baseline CE-CFM \cite{cheng2025c2ot}, the proposed method, and the ground-truth reference. From top to bottom, the Gaussian noise standard deviations are 0.22, 0.51, 0.61, and 1.00, respectively. The PSNR/SSIM values for each result are shown below the corresponding image.}
  \label{fig:voc}
\end{figure*}

Our experiments involve four datasets spanning three application scenarios: synthetic natural-image denoising, fluorescence microscopy denoising, and low-dose CT denoising.

\begin{enumerate}
  \item \textit{Synthetic natural-image denoising:} For natural-image experiments, we use the Pattern Analysis, Statistical Modelling and Computational Learning Visual Object Classes Challenge (PASCAL VOC)~\cite{Everingham2010VOC} dataset for training and validation, and evaluate generalization on the Berkeley Segmentation Dataset (BSDS~500)~\cite{arbelaez2011contour}. PASCAL VOC 2012 contains diverse natural scenes, textures, and objects, making it suitable for controlled noise injection during training. We adopt its standard training/validation split, using 11,530 images for training and 1,449 images for validation and ablation studies. For testing, BSDS~500 contains 500 natural images, and noisy observations are generated by adding zero-mean Gaussian noise with standard deviations uniformly sampled from \([0,1]\). In our formulation, \(x_0\) denotes the clean image, and the maximum noise level is normalized to 1, so that 
  \begin{equation}
      x_T=x_0+\epsilon,\qquad \epsilon\sim\mathcal{N}(0,I).
  \end{equation}
  This setting allows us to evaluate the model over a broad range of noise levels, from nearly clean inputs to the maximum noise level used during training.

  \item \textit{Fluorescence Microscopy Denoising Dataset (FMDD):} This dataset~\cite{Zhang2019FMD} contains 20 distinct samples. For each sample, it provides 50 noisy realizations at different averaging levels, including images averaged over 2, 4, 8, and 16 frames (avg2, avg4, avg8, avg16), together with a high-quality reference image obtained by averaging 50 frames. All images have a resolution of \(512\times512\). During training, all available noise levels are used as inputs, and the corresponding high-quality reference image is used as the target clean image \(x_0\). During testing, we evaluate the model separately at each noise level to assess its adaptability and robustness across different noise intensities.

  \item \textit{Mayo Clinic low-dose CT dataset:} This public dataset, released as part of the AAPM Low-Dose CT Grand Challenge~\cite{McCollough2017AAPM}, provides paired quarter-dose and normal-dose CT images. To evaluate denoising performance under controlled and adjustable noise levels, we further simulate the low-dose CT acquisition process based on clean baseline images. Specifically, a noisy CT image \(x_t\) is generated by combining Poisson noise in the photon domain with additive Gaussian noise in the image domain:
  \begin{equation}
      x_t=\mathcal{R}^{-1}\!\left(-\ln\left(\frac{I}{I_0}\right)\right)+\eta,\qquad
    I=\mathcal{P}\!\left(I_0 e^{-\mathcal{R}x_0}\right),
  \end{equation}
  where \(\mathcal{R}\) and \(\mathcal{R}^{-1}\) denote the Radon transform and the corresponding reconstruction operator, respectively, \(\mathcal{P}(\cdot)\) denotes Poisson sampling, \(I_0\) is the incident photon count, \(I\) is the transmitted photon count after attenuation, and \(\eta\sim\mathcal{N}(0,\sigma^2)\) denotes additive Gaussian noise in the image domain. By varying \(I_0\) and \(\sigma\), we control the noise intensity in the simulated low-dose CT images and evaluate the proposed method under a range of noise conditions that better reflect practical clinical scenarios.
\end{enumerate}

\subsection{Implementation Details}

The proposed model is implemented in Python using PyTorch. All experiments are conducted on a workstation equipped with an Intel Xeon Platinum 8358 CPU and 384 GB of RAM. Training is performed on four NVIDIA GeForce RTX 4090 GPUs, each with 24 GB of VRAM. We adopt a U-Net architecture as the backbone of the flow-matching network \(v_\theta\), which takes the current image state and the corresponding time step as input and predicts the vector field. The model is trained using the Adam optimizer with a learning rate of \(1\times10^{-4}\) and a batch size of 4 for 100 epochs. Validation performance is monitored during training to mitigate overfitting. To ensure that the model can handle a wide range of noise intensities, training noise levels are uniformly sampled from the same range used in testing.

During inference, we use a discrete time grid of length \(T=100\), where \(t_i=i/T\) for \(i=0,\dots,T-1\). The coarse sampling interval is set to \(M=10\), so that larger time steps are used in the early denoising stage corresponding to high-noise conditions, while finer steps are used near the clean endpoint for refined restoration. Before inference, the noise level of the input noisy image is first estimated, and the resulting estimate is then used to determine the starting time of reverse integration, as described in Section~\ref{sec:method}. Denoising performance is evaluated using standard metrics, including Peak Signal-to-Noise Ratio (PSNR) and Structural Similarity Index Measure (SSIM), with respect to the ground-truth clean images.

\subsection{Comparison With State-of-The-Art Methods}

\subsubsection{Results on Synthetic Noisy Natural Images}

\begin{figure}[ht]
  \centering
  \includegraphics[width=1.0\linewidth]{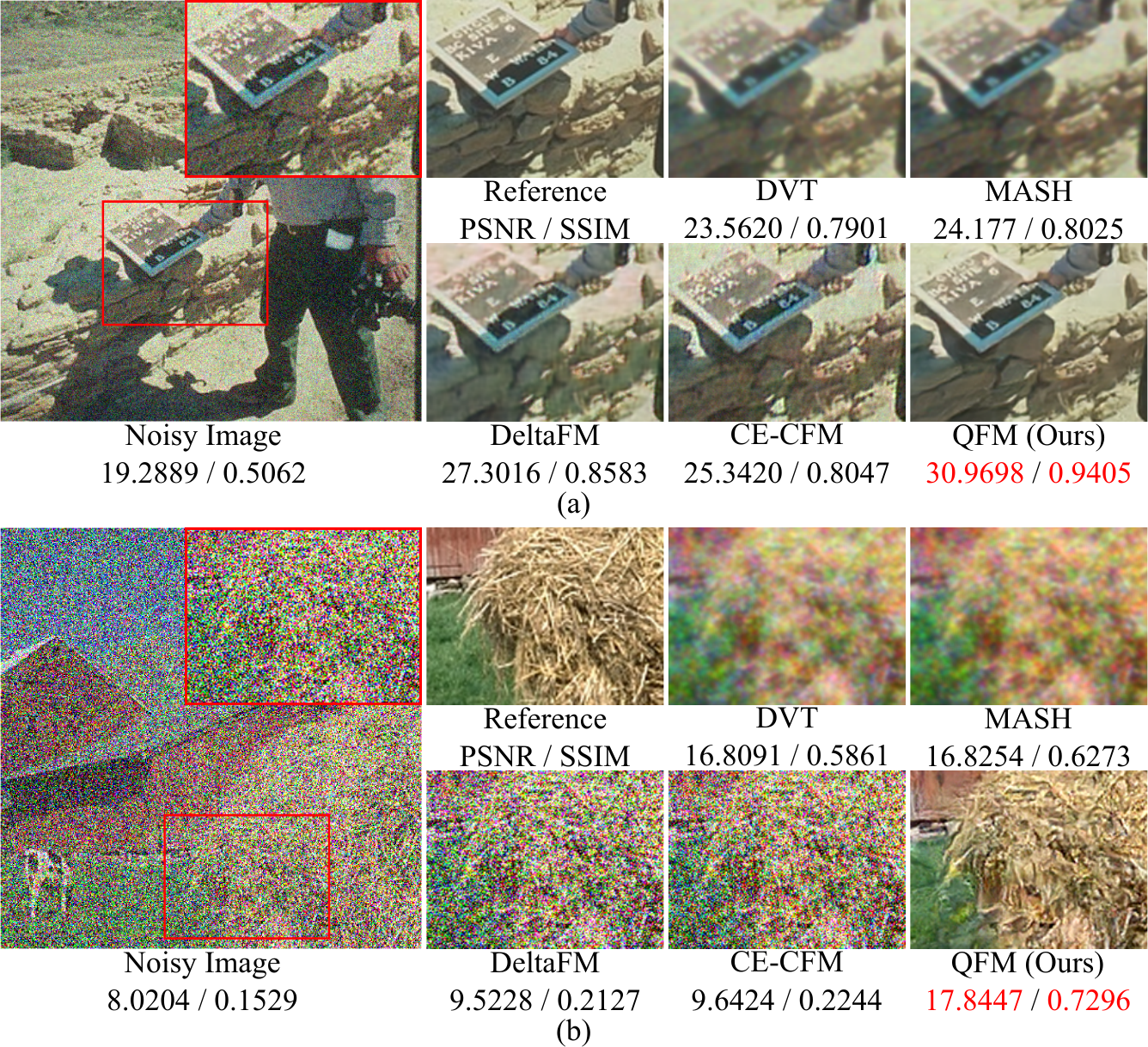}
  \caption{Qualitative comparison of restored ROIs under different noise levels. In each subfigure, the leftmost image shows the noisy input, where the red box indicates the selected ROI, and the remaining images show the corresponding magnified ROIs restored by different methods. (a) Low-noise case. (b) High-noise case. PSNR/SSIM values are reported below each result.}
  \label{fig:voc_roi}
\end{figure}

\begin{table*}[ht]
  \centering
  \caption{Comparison of PSNR (dB) and SSIM for different methods on the training, validation, and test sets. The medium-, high-, and very-high-noise results are obtained by further partitioning the test set according to the noise level.}
  \label{tab:radar_to_table}
  \setlength{\tabcolsep}{4pt}
  \begin{tabular}{llcccccccccccc}
    \toprule
    \toprule
    \multirow{3}{*}{\textbf{Method}} 
    & \multirow{3}{*}{\textbf{Venue}}
    & \multicolumn{6}{c}{\textbf{Overall splits}}
    & \multicolumn{6}{c}{\textbf{Test subsets by noise level}} \\
    \cmidrule(lr){3-8}
    \cmidrule(lr){9-14}
    &
    & \multicolumn{2}{c}{\textbf{Train}}
    & \multicolumn{2}{c}{\textbf{Validation}}
    & \multicolumn{2}{c}{\textbf{Test}}
    & \multicolumn{2}{c}{\textbf{Medium noise}}
    & \multicolumn{2}{c}{\textbf{High noise}}
    & \multicolumn{2}{c}{\textbf{Very high noise}} \\
    \cmidrule(lr){3-4}
    \cmidrule(lr){5-6}
    \cmidrule(lr){7-8}
    \cmidrule(lr){9-10}
    \cmidrule(lr){11-12}
    \cmidrule(lr){13-14}
    &
    & \textbf{PSNR} & \textbf{SSIM}
    & \textbf{PSNR} & \textbf{SSIM}
    & \textbf{PSNR} & \textbf{SSIM}
    & \textbf{PSNR} & \textbf{SSIM}
    & \textbf{PSNR} & \textbf{SSIM}
    & \textbf{PSNR} & \textbf{SSIM} \\
    \midrule
    Noisy input
    & --
    & 7.7660 & 0.0847
    & 7.5475 & 0.0895
    & 6.8602 & 0.0738
    & 22.7474 & 0.5974
    & 16.6710 & 0.3773
    & 5.0450 & 0.0158 \\
    DVT
    & ECCV~2024
    & 21.1297 & 0.5491
    & 21.2101 & 0.5483
    & 20.9563 & 0.5419
    & 24.6206 & 0.7986
    & 23.7915 & 0.7523
    & 20.4850 & 0.5078 \\
    MASH
    & CVPR~2024
    & 20.1750 & 0.5025
    & 20.2312 & 0.5149
    & 19.9320 & 0.5279
    & 25.7219 & 0.8232
    & 24.7795 & 0.7642
    & 19.1535 & 0.4892 \\
    DeltaFM
    & ICCV~2025
    & 15.1461 & 0.2522
    & 15.2940 & 0.2547
    & 14.5999 & 0.2296
    & 26.8789 & 0.8573
    & 27.2161 & 0.8193
    & 12.7341 & 0.1393 \\
    CE-CFM
    & NeurIPS~2025
    & 15.1861 & 0.2247
    & 15.0143 & 0.2118
    & 14.3655 & 0.1904
    & 26.9179 & 0.7830
    & 25.1600 & 0.6922
    & 12.6515 & 0.1102 \\
    \rowcolor{gray!20}
    \textbf{QFM}
    & Ours
    & \textbf{25.6718} & \textbf{0.6568}
    & \textbf{25.6962} & \textbf{0.6596}
    & \textbf{25.4821} & \textbf{0.6501}
    & \textbf{34.3397} & \textbf{0.9467}
    & \textbf{30.8646} & \textbf{0.9019}
    & \textbf{24.4781} & \textbf{0.6099} \\
    \bottomrule
    \bottomrule
  \end{tabular}
\end{table*}

To evaluate the proposed QFM method, we first conduct experiments on natural images corrupted by synthetic Gaussian noise and compare it with representative state-of-the-art methods from different paradigms. The baselines include DVT, a Transformer-based supervised method; MASH, a recent self-supervised approach; DeltaFM, a standard flow-matching model; and CE-CFM, a conditional flow-matching variant. Qualitative comparisons on the BSDS~500 test set under different noise levels are shown in Fig.~\ref{fig:voc}.

The supervised baseline DVT exhibits a reasonable denoising capability, but its performance degrades noticeably when the input noise level deviates from the training regime, often leading to oversmoothed textures and pixel-averaging artifacts. MASH shows relatively stronger adaptability across different noise levels, although its outputs may still contain residual noise and locally unstable structures. By contrast, the flow-based baselines DeltaFM and CE-CFM are more sensitive to the vector-field mismatch discussed in Section~\ref{sec:related_theory}, and their performance deteriorates more markedly under unseen noise levels. Although these models may achieve competitive results when training and testing are conducted under matched noise conditions, their generalization becomes clearly limited in the present cross-noise-level setting, as reflected by the remaining noise and visible artifacts in Fig.~\ref{fig:voc}.

A more detailed qualitative comparison is provided in Fig.~\ref{fig:voc_roi}, which focuses on representative ROIs under both low- and high-noise conditions. Under low noise, QFM preserves fine structures and local contrast more faithfully, whereas several competing methods exhibit varying degrees of oversmoothing. Under high noise, the advantage of QFM becomes more pronounced: it recovers the main structures of the ROI more reliably, while the other methods either fail to suppress the strong corruption adequately or produce heavily degraded textures. These enlarged comparisons further show that QFM achieves a better balance between noise suppression and detail preservation across substantially different noise levels.

For quantitative evaluation, Table~\ref{tab:radar_to_table} reports the PSNR and SSIM results on the training, validation, and test sets, while the medium-, high-, and very-high-noise results are obtained by further partitioning the test set according to the noise level. QFM ranks first on all three data splits in terms of both PSNR and SSIM, demonstrating consistently superior performance over all competing methods. Relative to the second-best baseline, QFM achieves PSNR improvements of 21.50\%, 21.15\%, and 21.60\% on the training, validation, and test sets, respectively, together with SSIM improvements of 19.61\%, 20.30\%, and 19.97\%. The consistency of these gains across different splits indicates that the proposed method generalizes robustly beyond a specific subset of the data.

More notably, the superiority of QFM remains evident across different noise regimes on the test set. Under medium noise, QFM surpasses the best competing baseline by 27.57\% in PSNR and 10.43\% in SSIM. Under high noise, the corresponding improvements are 13.41\% and 10.08\%, respectively. Even in the very-high-noise regime, where the restoration task becomes substantially more challenging, QFM still achieves gains of 19.49\% in PSNR and 20.11\% in SSIM over the best competing method. These results suggest that incorporating quantitative noise estimation into the inference process improves overall restoration quality and provides strong robustness across substantially different noise levels.

\subsubsection{Results on the FMDD Dataset}

\begin{figure*}[ht]
  \centering
  \includegraphics[width=0.95\linewidth]{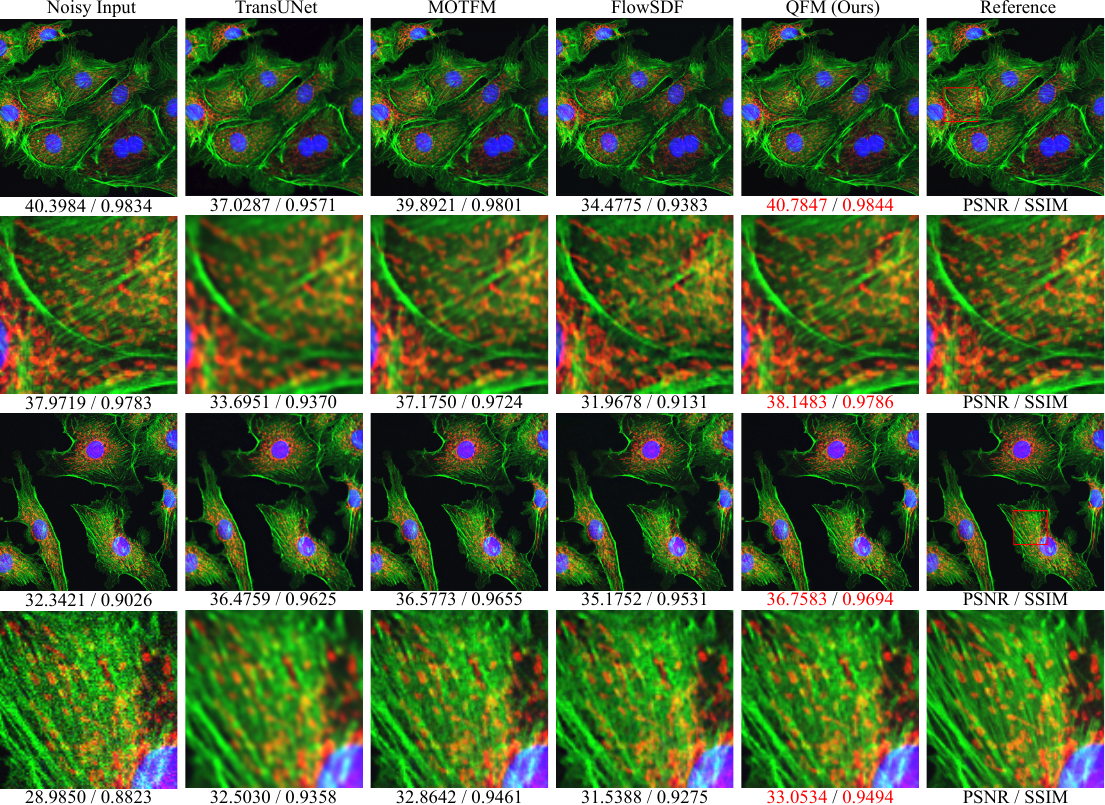}
  \caption{Qualitative comparison on the FMDD dataset. From left to right: noisy input, supervised baseline TransUNet \cite{chen2024transunet}, flow-based baseline MOTFM \cite{yazdani2025flow}, conditional flow-based baseline FlowSDF \cite{bogensperger2025flowsdf}, QFM (Ours), and the reference image. The second and fourth rows show enlarged ROI patches corresponding to the red boxes in the first and third rows, respectively. PSNR/SSIM values are reported below each result.}
  \label{fig:fmdd_results}
\end{figure*}

\begin{table*}[htbp]
  \centering
  \caption{Quantitative comparison of denoising performance on the FMDD dataset. Raw denotes the original noisy input, while Avg2, Avg4, Avg8, and Avg16 denote images obtained by averaging 2, 4, 8, and 16 frames, respectively.}
  \label{tab:FMDD_dataset}
  \setlength{\tabcolsep}{4pt}
  \begin{tabular}{llcccccccccc}
    \toprule
    \toprule
    \multirow{2}{*}{\textbf{Method}} & \multirow{2}{*}{\textbf{Venue}}
    & \multicolumn{2}{c}{\textbf{Raw}}
    & \multicolumn{2}{c}{\textbf{Avg2}}
    & \multicolumn{2}{c}{\textbf{Avg4}}
    & \multicolumn{2}{c}{\textbf{Avg8}}
    & \multicolumn{2}{c}{\textbf{Avg16}} \\
    \cmidrule(lr){3-4}
    \cmidrule(lr){5-6}
    \cmidrule(lr){7-8}
    \cmidrule(lr){9-10}
    \cmidrule(lr){11-12}
    &
    & \textbf{PSNR} & \textbf{SSIM}
    & \textbf{PSNR} & \textbf{SSIM}
    & \textbf{PSNR} & \textbf{SSIM}
    & \textbf{PSNR} & \textbf{SSIM}
    & \textbf{PSNR} & \textbf{SSIM} \\
    \midrule
    Noisy Image
    & --
    & 27.9524 & 0.5735
    & 30.8260 & 0.7003
    & 33.6758 & 0.8114
    & 36.6080 & 0.8947
    & 40.1838 & 0.9501 \\
    TransUNet
    & MIA~2024
    & 34.7428 & 0.8983
    & 35.5516 & 0.9123
    & 36.0403 & 0.9192
    & 36.4022 & 0.9228
    & 36.8469 & 0.9249 \\
    MOTFM
    & MICCAI~2025
    & 31.9868 & 0.7528
    & 34.7947 & 0.8546
    & 37.1023 & 0.9160
    & 39.0293 & 0.9489
    & 41.0715 & 0.9661 \\
    FlowSDF
    & IJCV~2025
    & 31.9759 & 0.7788
    & 33.4901 & 0.8529
    & 34.1086 & 0.8835
    & 34.3907 & 0.8973
    & 34.6921 & 0.9054 \\
    \rowcolor{gray!20}
    \textbf{QFM}
    & Ours
    & \textbf{35.0272} & \textbf{0.9040}
    & \textbf{36.5377} & \textbf{0.9264}
    & \textbf{38.0481} & \textbf{0.9444}
    & \textbf{39.6858} & \textbf{0.9595}
    & \textbf{41.9457} & \textbf{0.9721} \\
    \bottomrule
    \bottomrule
  \end{tabular}
\end{table*}

\begin{figure*}[htbp]
  \centering
  \includegraphics[width=0.95\linewidth]{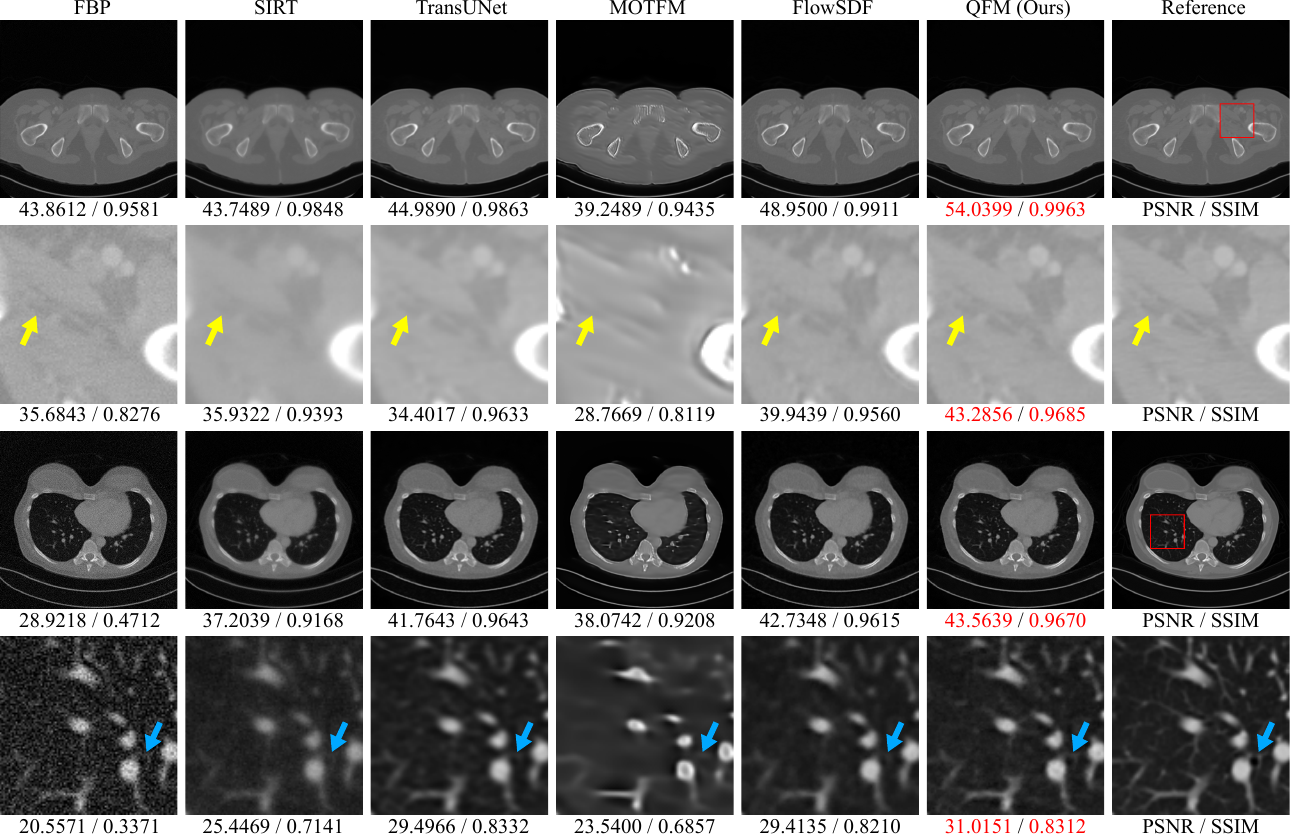}
  \caption{Qualitative comparison on the Mayo dataset. From left to right: FBP, SIRT, TransUNet, MOTFM, FlowSDF, QFM (Ours), and the reference image. The second and fourth rows show enlarged ROI patches corresponding to the red boxes in the first and third rows, respectively. Display windows are set to [-1000~HU, 1450~HU] for the full images and [-1000~HU, 630~HU] for the ROI patches. PSNR/SSIM values are reported below each result.}
  \label{fig:med_results}
\end{figure*}

\begin{table*}[htbp]
  \centering
  \caption{Quantitative comparison of denoising performance on the Mayo dataset.}
  \label{tab:mayo_dataset}
  \begin{tabular}{cccccccc}
    \toprule
    \toprule
    \textbf{Anatomical Region} & \textbf{Number of Images} & \textbf{Metric} & \textbf{Noisy Image} & \textbf{TransUNet} & \textbf{MOTFM} & \textbf{FlowSDF} & \cellcolor{gray!20}\textbf{QFM (Ours)} \\
    \midrule
    \multirow{2}{*}{\centering Chest}
    & \multirow{2}{*}{\centering 1047}
    & PSNR & 28.7306 & 42.7329 & 40.7846 & 42.9543 & \cellcolor{gray!20}\textbf{44.2525} \\
    & & SSIM & 0.4577 & 0.9505 & 0.9416 & 0.9477 & \cellcolor{gray!20}\textbf{0.9649} \\
    \midrule
    \multirow{2}{*}{\centering Abdomen}
    & \multirow{2}{*}{\centering 1205}
    & PSNR & 27.7540 & 41.8156 & 39.1901 & 42.8373 & \cellcolor{gray!20}\textbf{44.0748} \\
    & & SSIM & 0.4428 & 0.9476 & 0.9402 & 0.9493 & \cellcolor{gray!20}\textbf{0.9591} \\
    \bottomrule
    \bottomrule
  \end{tabular}
\end{table*}

To further evaluate the generalization capability of the proposed framework under real and complex noise conditions, we extend our experiments to the FMDD dataset and the Mayo low-dose CT dataset. It should be noted that the noise in these real datasets typically consists of both Gaussian and Poisson components. Since many self-supervised denoising methods rely on a zero-mean noise assumption, their modeling assumptions are not fully consistent with this setting. Our preliminary experiments also indicated relatively weak performance of such methods in this scenario, and they were therefore not included in the final comparison. Instead, we compare with representative supervised and generative baselines in medical image analysis, including TransUNet, MOTFM, and FlowSDF. The visual results are shown in Fig.~\ref{fig:fmdd_results}.

Because some subsets have relatively high signal-to-noise ratios, the differences among methods are not always obvious at the full-image scale. However, the magnified ROIs in Fig.~\ref{fig:fmdd_results} reveal clearer distinctions. TransUNet tends to produce oversmoothed structures in some regions, leading to the loss of fine textures. MOTFM preserves more details than the supervised baseline, but its results are still slightly less faithful than those of QFM in local structures. FlowSDF shows comparatively weaker restoration quality in both examples. By contrast, QFM better preserves fine filamentary structures and local contrast, and its restored ROIs are visually closer to the reference images.

A more conclusive comparison is provided by the quantitative results in Table~\ref{tab:FMDD_dataset}. The compared methods generally face greater difficulty on the Raw subset and, to a lesser extent, on Avg2, which correspond to relatively high-noise conditions. As the averaging number increases and the noise level decreases, the restoration quality of all methods improves. Nevertheless, QFM remains consistently superior across all subsets. It achieves the best performance on the high-noise subsets (Raw and Avg2) and continues to maintain the highest image quality on the relatively cleaner subsets (Avg4, Avg8, and Avg16), indicating strong robustness across a broad range of real noise levels.

\subsubsection{Results on the Mayo Low-Dose CT Dataset}

On the Mayo low-dose CT dataset, we evaluate the proposed method under mixed Poisson--Gaussian noise over a broader intensity range. As shown in Fig.~\ref{fig:med_results}, QFM preserves subtle anatomical structures more faithfully than the competing methods. In the enlarged ROIs, the proposed method recovers weak tissue boundaries and fine local textures with clearer definition and closer visual agreement to the reference image. By contrast, FBP and SIRT contain stronger residual noise and reconstruction artifacts, while the learning-based baselines tend to lose part of the weak structural details or introduce excessive smoothing in local regions.

A closer inspection of the ROIs in Fig.~\ref{fig:med_results} further highlights the advantage of the proposed method. In the second row, the yellow arrow indicates a low-contrast soft-tissue region. This area is particularly challenging because the boundary cue is weak and can be easily obscured by noise or oversmoothing. Compared with the competing methods, QFM preserves the subtle intensity transition and local structural continuity more faithfully, yielding a clearer and more reference-consistent depiction of the low-contrast region. By contrast, FBP and SIRT still contain noticeable noise fluctuations, TransUNet tends to oversmooth the weak boundary, and MOTFM introduces visible distortion in the local structure. In the fourth row, the blue arrow highlights a small alveolar air-space pattern in the lung parenchyma. This fine structure is highly sensitive to both residual noise and excessive smoothing. QFM reconstructs these small hole-like structures with better separation, sharper boundaries, and more faithful morphology, whereas the other methods either blur the individual air spaces together or fail to suppress the background corruption sufficiently. These ROI comparisons suggest that QFM is better able to recover weak low-contrast structures and delicate pulmonary textures simultaneously, which is particularly important for preserving diagnostically relevant anatomical details in low-dose CT images.

The quantitative results in Table~\ref{tab:mayo_dataset} further support these observations. QFM achieves the best performance in both anatomical regions, reaching 44.2525 dB / 0.9649 on the chest subset and 44.0748 dB / 0.9591 on the abdomen subset. These results consistently outperform the competing baselines, indicating that the proposed framework generalizes well across different anatomical regions under complex low-dose CT noise. This advantage is consistent with the core design of QFM: by adapting the inference trajectory according to the estimated input noise level, the model remains more robust when the underlying noise characteristics vary across samples.

Taken together, the results on FMDD and the Mayo low-dose CT dataset support the main conclusion of this work: by quantitatively estimating the input noise level and adapting the reverse generative trajectory accordingly, QFM maintains strong generalization across a continuous range of real and complex noise conditions. This behavior is consistent on both microscopy and CT data, indicating the practical value of the proposed framework beyond synthetic settings.

\subsection{Ablation Study}

This subsection analyzes the contribution of quantitative noise estimation to the inference process of the proposed framework. The accuracy of the proposed noise-level estimation method is evaluated separately. Here, we focus on its effect on adaptive flow-matching inference.

Specifically, we compare the proposed QFM framework with an ablation variant that does not use quantitative noise estimation to adapt the inference process. In this non-adaptive variant, the reverse denoising trajectory is performed with a fixed inference configuration regardless of the actual input noise level. By contrast, QFM determines the starting point and the step-size schedule according to the estimated noise level. This comparison is designed to evaluate whether noise-aware adaptive inference is necessary for robust denoising across varying noise conditions.

\begin{figure}[ht]
  \centering
  \includegraphics[width=1.0\linewidth]{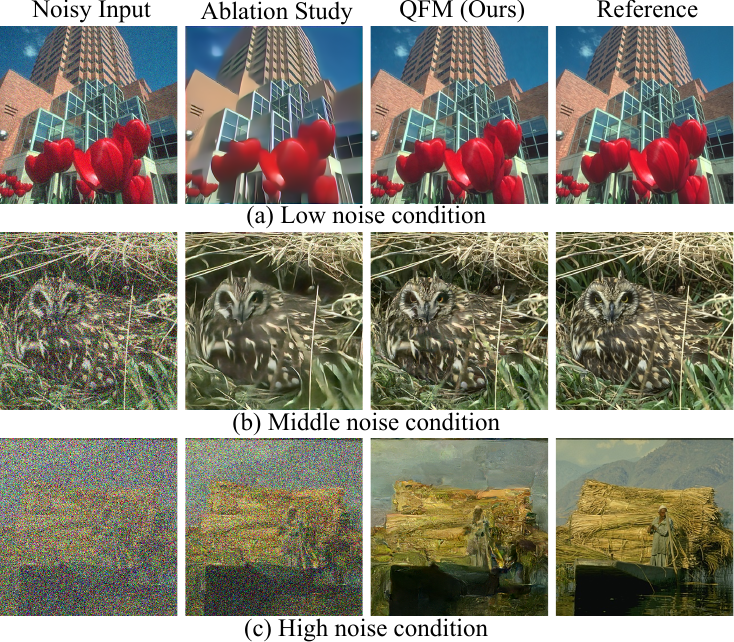}
  \caption{Qualitative comparison of ablation results under different noise conditions. From left to right: noisy input, the variant without quantitative noise estimation, QFM (Ours), and the reference image. (a) Low-noise condition. (b) Medium-noise condition. (c) High-noise condition.}
  \label{fig:show_ablation}
\end{figure}

\begin{figure}[ht]
  \centering
  \includegraphics[width=1.0\linewidth]{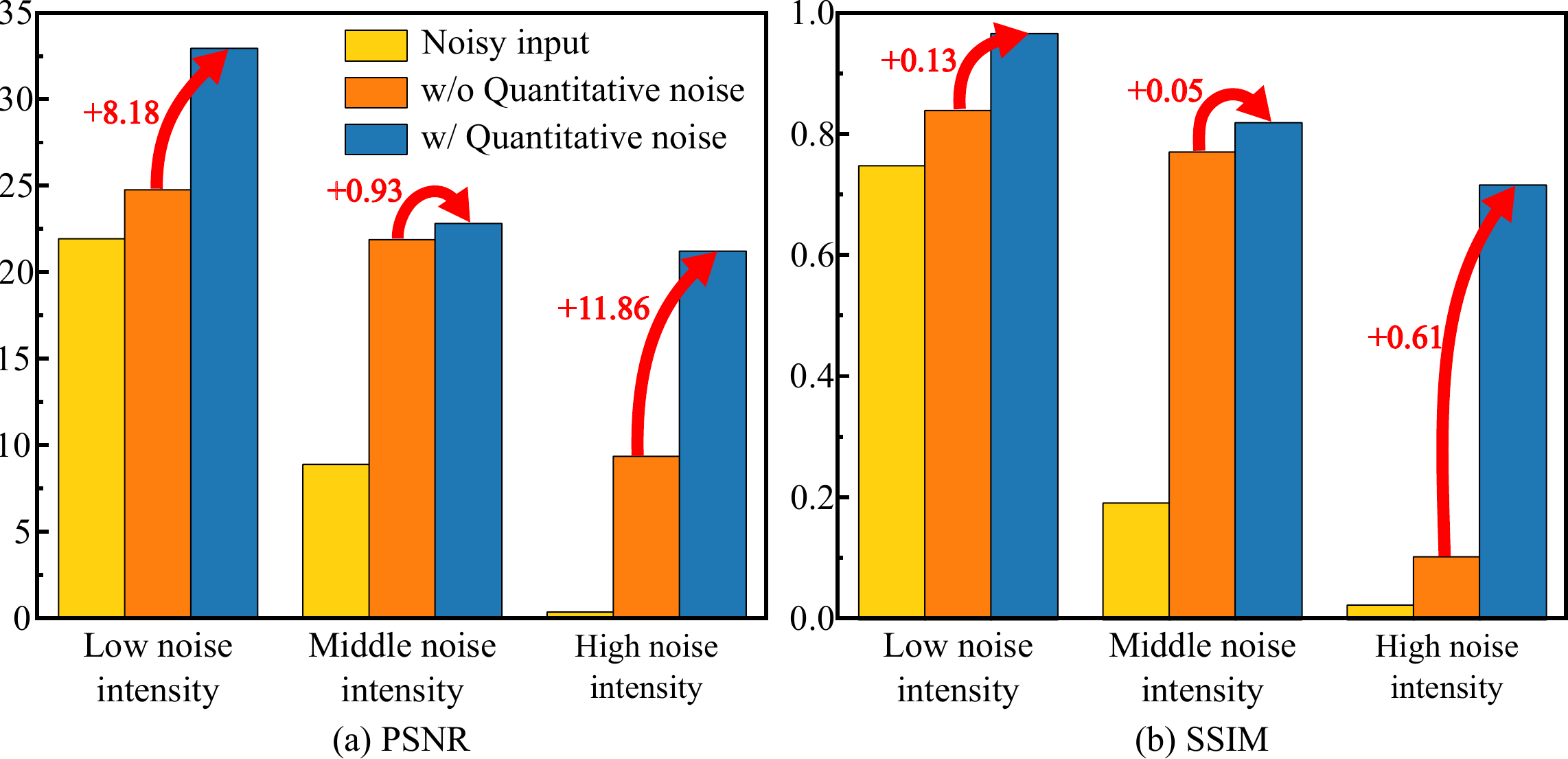}
  \caption{Quantitative comparison of the ablation study under different noise intensities. Compared with the variant without quantitative noise estimation, QFM consistently improves both PSNR and SSIM, with the largest gains observed under high-noise conditions.}
  \label{fig:hist_psnr_ssim}
\end{figure}

\begin{figure*}[ht]
  \centering
  \includegraphics[width=1.0\linewidth]{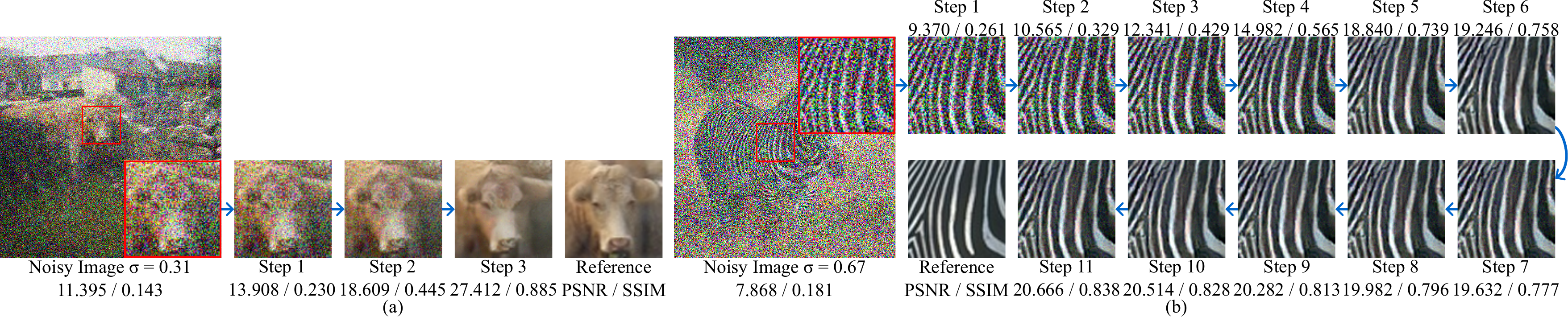}
  \caption{Visualization of the progressive denoising process under different noise levels. Different input noise intensities lead to different effective reverse trajectories and different numbers of denoising steps. (a) Low-noise case with \(\sigma=0.31\). (b) High-noise case with \(\sigma=0.67\).}
  \label{fig:traj}
\end{figure*}

As shown in Fig.~\ref{fig:show_ablation}, the non-adaptive variant exhibits clear limitations when the input noise level differs from the nominal inference setting. Under low-noise conditions, the reverse denoising process becomes unnecessarily aggressive, which leads to oversmoothed results and the loss of fine textures. This can be observed in Fig.~\ref{fig:show_ablation}(a), where local structural details become noticeably blurred. When the input noise level is closer to the nominal setting, the performance gap becomes smaller, as shown in Fig.~\ref{fig:show_ablation}(b). Under high-noise conditions, however, the fixed inference configuration becomes insufficient for adequate denoising, leaving strong residual noise and degraded structures in the output, whereas QFM is able to recover the main image content much more effectively, as shown in Fig.~\ref{fig:show_ablation}(c).

The quantitative results in Fig.~\ref{fig:hist_psnr_ssim} are consistent with these observations. Compared with the non-adaptive variant, QFM improves PSNR by 11.86 dB, 0.93 dB, and 8.18 dB under high-, medium-, and low-noise conditions, respectively, while the corresponding SSIM gains are 0.61, 0.05, and 0.13. These results indicate that quantitative noise estimation is particularly beneficial when the input noise level deviates substantially from the nominal inference setting, and that it leads to more stable denoising performance across the full noise range.

In addition, Fig.~\ref{fig:traj} visualizes the progressive inference process under different noise levels. For a relatively low-noise input (\(\sigma=0.31\)), the image quality improves rapidly within only a few reverse steps, and the restored result becomes visually close to the reference after three steps. For a higher-noise input (\(\sigma=0.67\)), more reverse steps are required, and both PSNR and SSIM continue to improve as the denoising trajectory progresses. This behavior directly illustrates the rationale of the proposed framework: different input noise levels should correspond to different effective inference trajectories. The results therefore confirm the necessity and effectiveness of quantitative noise estimation for adaptive cross-noise-level denoising.

\section{Conclusion}
\label{sec:conclusion}

This paper investigates the limitations of existing flow-matching-based denoising methods under varying noise levels and, on this basis, proposes \emph{Quantitative Flow Matching} (QFM). The central idea of QFM is to estimate the input noise level quantitatively from local pixel statistics and to use this estimate to adapt the reverse inference process, including the starting point on the normalized trajectory and the integration schedule. In this way, the restoration process is matched more closely to the actual noise intensity of the input, rather than relying on a fixed inference configuration. Extensive experiments on natural-image, medical CT, and microscopy datasets show that QFM achieves strong and stable denoising performance across different noise types and noise levels, while also reducing redundant computation.

Despite these advantages, the proposed method still has several limitations. First, it relies on a global noise-level estimate and therefore assumes that the dominant noise level is sufficiently representative of the entire image. This assumption may not hold for spatially non-uniform or highly signal-dependent noise. Second, handling strongly non-stationary noise may require a more refined spatially adaptive formulation, for example by designing region-wise or pixel-wise inference trajectories, which remains beyond the scope of the current work. Third, the performance of the framework still depends on the quality of the pre-trained flow-matching model and the adopted trajectory parameterization. Inaccuracies in these components may affect the final restoration quality.

Overall, this work provides a quantitative and adaptive inference framework for applying generative models to practical image denoising tasks. By explicitly estimating the input noise level and using it to configure the reverse denoising trajectory, QFM alleviates an important limitation of existing methods when handling unknown and varying noise conditions. We expect that this idea may further motivate the development of more robust, efficient, and practically applicable generative methods for image restoration.

\section*{Acknowledgments}
The authors are grateful to the National Center for Applied Mathematics Beijing for funding this research work, and to the Beijing Higher Institution Engineering Research Center of Testing and Imaging for its support and assistance during this study.

\bibliographystyle{IEEEtran}
\bibliography{Bibliography}

\vspace{11pt}

\vfill

\end{document}